\documentclass[letterpaper, 10 pt, conference]{ieeeconf}  

\IEEEoverridecommandlockouts                              

\usepackage{amsmath} 
\usepackage{cases}
\usepackage{amssymb}  
\usepackage[colorinlistoftodos]{todonotes}
\usepackage{hyperref}
\usepackage{multirow}
\usepackage{booktabs}
\usepackage[colorinlistoftodos]{todonotes}
\usepackage{textcomp}
\usepackage{gensymb}
\usepackage{siunitx}
\usepackage{url}
\usepackage{mathtools}
\usepackage{subcaption}
\usepackage{bm}  
\usepackage{soul}
\usepackage{dblfloatfix}
\usepackage{cite}

\providecommand{\keywords}[1]{\textbf{\textit{Index terms---}} #1}

\usepackage{accents}

\newcommand{\vlambda}{\mbox{\boldmath $\lambda$}}

\newcommand{\vpi}{\mbox{\boldmath $\pi$}}

\newcommand{\vtau}{\mbox{\boldmath $\tau$}}

\newcommand{\vch}{\mbox{\boldmath $\chi$}}

\newcommand{\vf}{\bm f}

\newcommand{\vh}{\bm h}

\newcommand{\vn}{\bm n}

\newcommand{\vq}{\bm q}

\newcommand{\vu}{\bm u}
\newcommand{\vv}{\bm v}

\newcommand{\vx}{\bm x}

\newcommand{\vB}{\bm B}

\newcommand{\vD}{\bm D}

\newcommand{\vH}{\bm H}
\newcommand{\vI}{\bm I}
\newcommand{\vJ}{\bm J}
\newcommand{\vK}{\bm K}

\newcommand{\vM}{\bm M}

\newcommand{\vQ}{\bm Q}
\newcommand{\vR}{\bm R}
\newcommand{\vS}{\bm S}

\newcommand{\vY}{\bm Y}

\title{\LARGE \bf Model Predictive Robot-Environment Interaction Control for Mobile Manipulation Tasks}

\author{Maria Vittoria Minniti, Ruben Grandia, Kevin Fäh, Farbod Farshidian, Marco Hutter 
\thanks{This work was supported in part by the Swiss National Science Foundation through the National Centre of Competence in Research Robotics (NCCR Robotics), in part by the Swiss National Science Foundation through the National Centre of Competence in Digital Fabrication (NCCR dfab), in part by Intel Network on Intelligent Systems, and in part by the European Union’s Horizon 2020 research and innovation programme under grant agreement No 780883.)} 
\thanks{All authors are with Robotic Systems Lab, ETH Zurich, Zurich 8092, Switzerland {\tt\footnotesize \{mminniti, rgrandia, farbodf, mahutter\}@ethz.ch}}%
}

\begin{document}

\maketitle
\thispagestyle{empty}
\pagestyle{empty}


\begin{abstract}
Modern, torque-controlled service robots can regulate contact forces when interacting with their environment. Model Predictive Control (MPC) is a powerful method to solve the underlying control problem, allowing to plan for whole-body motions while including different constraints imposed by the robot dynamics or its environment. However, an accurate model of the robot-environment is needed to achieve a satisfying closed-loop performance. Currently, this necessity undermines the performance and generality of MPC in manipulation tasks. In this work, we combine an MPC-based whole-body controller with two adaptive schemes, derived from online system identification and adaptive control. 
As a result, we enable a general mobile manipulator to interact with unknown environments, without any need for re-tuning parameters or pre-modeling the interacting objects. 
In combination with the MPC controller, the two adaptive approaches are validated and benchmarked with a ball-balancing manipulator in door opening and object lifting tasks.
\end{abstract}


\section{INTRODUCTION}

Mobile service robots deployed in human habitats will face a wide variety of manipulation tasks. Assisted lifting of objects, cleaning of windows, or opening of doors, all require deliberate interaction between the robot and its surroundings. Solving such robot-environment interaction control tasks in a stable and repeatable manner is still an open problem. For fixed-base robots, state-of-the-art control schemes \cite{siciliano2010robotics}, \cite{Natale2003interaction}, \cite{Vukobratovic2009} can define and limit the interaction forces in a reactive manner. However, if the manipulator is mounted on a mobile base, control action is additionally demanded for the stabilization of the robotic platform itself. This is especially the case for underactuated mobile bases, such as humanoids, quadrupeds or wheeled balancing systems, that require a planning strategy to maintain their balance \cite{koenemann2015whole, sleiman2021unified, gupta20192d}. For known environments \cite{9197524} or simple interactions~\cite{minniti2019whole}, planning for the robot motion and interaction has been addressed by a Model Predictive Controller (MPC), which deals with the constraints and dynamics by solving an optimal control problem in a receding horizon fashion. 

\begin{figure}[htp]
   \centering
   \includegraphics[scale=0.13]{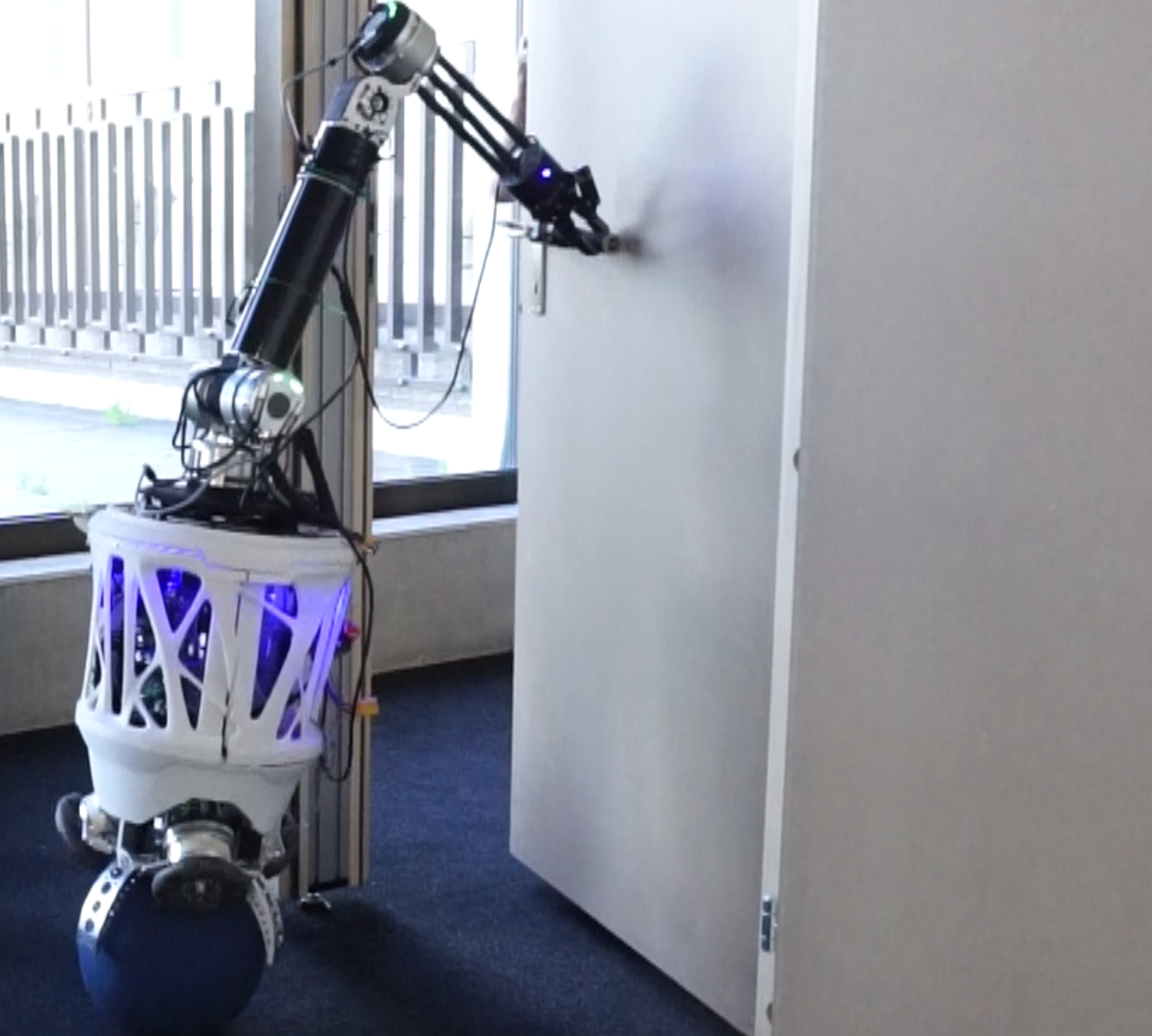}
   \caption{A ball-balancing manipulator opening a hinged door. The robot is required to maintain its balance and, at the same time, follow a desired end-effector trajectory while interacting with an unknown, external environment.}    
\label{fig:ballbot}    
\end{figure}

While robust to certain model mismatches, the lack of a precise system model can profoundly affect the MPC's performance. 
For interaction tasks, this includes a model of the environment that the robot should interact with, which, in general, is not available and needs to be extracted in advance.
To tackle this issue, we extend an SLQ-MPC framework introduced in~\cite{Farshidian2017MPC} by two adaptive control strategies, that are based on model identification adaptive control (MIAC) \cite{oreg2019model} and model reference adaptive control (MRAC) \cite{nguyen2018model}, \cite{siciliano2010robotics}. 
The design of the proposed methods relies on the assumption that the environment behaves as a linear mechanical system along the interaction trajectory direction. We benchmark the performance of these two adaptive approaches and the baseline non-adaptive MPC for opening a door and lifting a weight. Such applications are typically used to evaluate robotic manipulation skills \cite{quispe2018taxonomy}. We then conclude that while both techniques improve the MPC's performance during the interaction, the MRAC method demonstrates a more reliable and robust performance across the two different tasks.

For the experiments, a \emph{ballbot} with an arm has been employed, as shown in Fig.~\ref{fig:ballbot}. Due to its underactuation, a ballbot requires a whole-body control strategy that coordinates the movement and balance of the manipulator. Thus, controlling such a platform is an intriguing problem, that has received growing attention in the recent years \cite{li2019toward}, \cite{sonnleitner2019mechanics}, \cite{shomin2015sit}.

\subsection{Related Work} \label{related_work}

Classical methods in robot-environment interaction control are based on direct and indirect force control. In direct force control \cite{siciliano2010robotics}, the manipulation task is performed by tracking a desired force. This method is effective in all those scenarios where specific force profiles have to be applied. However, the generation of suitable force trajectories is not trivial without prior knowledge about the environment and the task. Indirect force control allows for tracking a desired force by exploiting its relation with deviations in position and can be formulated as an impedance or admittance controller \cite{hogan1985impedance}, \cite{kazerooni1986robust}. 
Impedance control allows tuning the controller's stiffness and damping based on the desired task, but its performance degrades when the environment's model is unknown or time-varying. Recent approaches propose to learn an error force term to compensate for the uncertainties in the robot and the environment model \cite{amanhoud2020force}. Another possibility is to change the controller gains based on the force tracking error \cite{lee2008force}, or an online estimate of the environment model \cite{diolaiti2005contact},\cite{mallapragada2007new}. The MIAC method presented in this letter fits in this category, with the novelty of using the environment estimates to improve the internal model of the MPC planner in real-time.

On the other hand, adaptive controllers have proven to be very effective when dealing with the uncertainty of the inertial parameters in the dynamic model of robotic manipulators \cite{siciliano2010robotics}, \cite{slotine1987adaptive}. In such methods, the adaptive parameters do not necessarily converge to the real values, but are updated to achieve convergence of the tracking error. Analysis on adaptive control has also been conducted in the context of force/position control in unknown interactions \cite{villani1999exponentially}, \cite{roy2002adaptive}, and it has mainly considered fixed-base, fully-actuated manipulators. The MRAC controller presented in this work focuses on obtaining an adaption law for the environment model, to be set as a constraint in an MPC module that can plan for motion and interaction force for a general mobile manipulator system. There are previous works that have used an adaptive law to update a parametric MPC model \cite{zhu2016lyapunov, zhu2019constrained}. However, so far this combination has not been addressed in a mobile manipulation scenario, especially when dealing with underactuated platforms.

\subsection{Contributions}

The main contribution of this work is to enable an MPC strategy to plan and control mobile-manipulation tasks that involve interaction with an unknown environment, without the need for re-tuning the control parameters for different tasks or offline modeling. To this end, we propose to combine a whole-body MPC scheme with two adaptive control techniques, based on system identification and model reference adaptive control. We benchmark the performance of the two described methods in two different representative manipulation tasks. We experimentally demonstrate that the combination of MPC and the model reference adaptive controller enables the robot to complete the desired tasks, while the performance of the baseline controller degrades, and the MPC scheme relying on online system identification fails to generalize across different tasks.

\section{MPC PROBLEM} \label{mcp_control}
In our proposed method, an MPC solver manages the online switching between subsystems (e.g., the change between cost functions and constraints) and the system model update at the start of every MPC iteration. 
We employ an SLQ-MPC method \cite{farshidian2017real}, which solves the underlying optimal control problem using the Sequential Linear Quadratic (SLQ) algorithm \cite{farshidian2017sequential}, while allowing to switch between subsystems.

Let $\vq, \dot \vq \in \mathbb{R}^n$ be the generalized positions and velocities of the robot, and let $\vtau \in \mathbb{R}^m$ be the actuation torques. In addition, the end-effector pose is defined as $\vch_{ee} \in SE(3)$.
The employed non-linear optimal control formulation aims at minimizing the cost function 
\begin{align}
    J^*(\vx_{ss}(0)) =& 
    \min_{\vu(\cdot)} \int_{0}^{T} ||\vx_{ss}^{d} - \vx_{ss} ||_{\vQ_{ss}}^{2} + ||\vu^{d} - \vu||_{\vR}^{2} \notag
    \\
    &\phantom{\min \int} + ||\vch_{ee}^{d} \boxminus  \vch_{ee}||_{\vQ_{ee}}^{2} \ \text{d}t,
\label{eq:cost_function}
\end{align}
where the state and input at time $t$ are defined as $\vx_{ss} := (\vq, \dot \vq) \in \mathbb{R}^{2n}, \vu:=\vtau \in \mathbb{R}^m$, and we use the symbol $\boxminus$ to denote the task-space pose error, as defined in \cite{siciliano2010robotics} for the quaternion rotation part. The optimization is subject to a set of constraints:
\begin{align}
    & {\vx_{ss}}(0) = \vx_{{ss}_0} \label{eq:initial_conditions}\\
    & \dot \vx_{ss} = \vf(\vx_{ss}, \vu) \label{eq:system_dynamics}\\
    & \vh(\vx_{ss}, \vu, t) \geq \mathbf{0}. \label{eq:ineq_constraints}
\end{align}
The initial condition of the MPC problem \eqref{eq:initial_conditions} coincides with the robot measured state. The system in~\eqref{eq:system_dynamics} consists of the robot's dynamical model and the interaction model, which are discussed in the following section.
Inequality constraints \eqref{eq:ineq_constraints} include joint angle and torque limits, and are treated as soft constraints according to a relaxed log-barrier function method \cite{grandia2019feedback}.

\section{INTERACTION CONTROL} \label{sec:interaction_control}
\subsection{Control Task}
\label{subsec:control_task}
The control task considered in this work is to follow a trajectory with the robot end-effector while having contact with an unknown environment. 
For trajectory generation, we use the time-optimal algorithm described in \cite{ramos2013time}, which allows for generating position and velocity trajectories for the end-effector with bounded acceleration and velocity magnitudes. Its simple implementation also allows fast recomputation. This is important for the door opening task, where the door hinge position and current angle are estimated online and thus the trajectory needs to be continuously updated.  

\subsection{Assumptions and Notation} \label{subsec:assumptions}
Let $\vlambda_{ee} \in \mathbb{R}^3$ be the interaction force between the robot end-effector and the environment, and let $\vv(t) \in \mathbb{R}^3$ be the unit vector tangent to the manipulation trajectory. Without loss of generality, we suppose $\vlambda_{ee} = \lambda \vv$, with $\lambda \in \mathbb{R}$; possible normal components could be compensated using force feedback, without modifications to the following derivation.

We state the following assumptions:
\begin{itemize}
\item
The contact between the end-effector and the environment is rigid.
\item The environment model can be described by a 3D second-order system:
\begin{equation}
    \vlambda_{ee} = \bar \vM \ddot \vx_{\text{env}} + \bar \vB  \dot \vx_{\text{env}} + \bar \vK (\vx_{\text{env}} - \vx_0) + \vf_s,
    \label{3d_mass_spring_damper}
\end{equation}
where $\vx_{\text{env}} \in \mathbb{R}^3$ is the position of the environment interaction point, $\bar \vM, \bar \vB, \bar \vK \in \mathbb{R}^{3 \times 3}$ are positive symmetric impedance matrices, $\vf_s \in \mathbb{R}^3$ is a static force, and $\vx_0 \in \mathbb{R}^3$ is a given spring equilibrium position.
\item The manipulation direction $\vv$ is a principal direction for $\bar \vM, \bar \vB, \bar \vK$, with $m, b, k \in \mathbb{R}$ as associated eigenvalues. These are the mass, damping and stiffness of the environment object.
\end{itemize}
 The model in Eq.~\eqref{3d_mass_spring_damper} applies to a number of interaction objects used in service robotics tasks. For instance, doors and windows usually have a spring and damping behavior. Carts and payloads can also be described by Eq.~\eqref{3d_mass_spring_damper}, setting $\bar \vB=\bm{0}$, $\bar \vK=\bm{0}$. The stated assumptions yield:
\begin{equation*}
    \lambda = \vv^T \vlambda_{ee} = m \vv^T \ddot \vx_{ee} + b\vv^T\dot \vx_{ee} + k \vv^T(\vx_{ee} - \vx_0) + \vv^T \vf_s,
\end{equation*}
where $\vx_{ee} = \vx_{\text{env}}$, since the contact is rigid. To derive a controller for the environment system, this equation is written as a one-dimensional second-order linear system. Many state space representations are possible, depending on the trajectory $\vv$. One possibility that allows to generalize among many trajectories consists in letting $x:=\vv^T \vx_{ee}$, which yields:
\begin{equation}
    \lambda = m\ddot x + b \dot x + k(x-x_0)+f_s + \mathcal{O}(\dot \vv),
\end{equation}
where $f_s = \vv^T \vf_s$ is the projected static force. We now assume the direction $\vv(t)$ to be constant or slowly varying; this applies to a number of relevant tasks, which have usually linear or circular paths on relatively small angles.
Thus, we obtain a robot-environment model as
\begin{subequations}
\begin{align}
    &\vM(\vq)\ddot \vq +\vn(\vq, \dot \vq) = \vS^{T}\vtau - \vJ_{ee}(\vq)^{T}(\vf_{ee} + \vlambda_{ee}),
    \label{eq:robot_eom}
    \\
    &m \ddot x + b \dot x + k(x-x_0)+f_s = \lambda,
    \label{eq:environment_eom}
\end{align}
\label{eq:robot_environment_eom}
\end{subequations}
where $\vM \in \mathbb{R}^{n \times n}$, and $\vn \in \mathbb{R}^n$ are the robot's mass matrix and non-linear terms, respectively. $\vS^T \in \mathbb{R}^{n \times m}$ maps the actuator torques to generalized torques and, in the case of a ballbot system, is a non-linear function of the joint positions $\vq$. $\vJ_{ee}(\vq) \in \mathbb{R}^{3 \times n}$ is the position jacobian of the end-effector frame. Since $x$ is a function of $\vq$,~\eqref{eq:environment_eom} can be written in terms of $\vq, \dot \vq, \ddot \vq$. Substituting $\vlambda_{ee} = \lambda \vv$ from~\eqref{eq:environment_eom}  into~\eqref{eq:robot_eom} and converting to state-space form, the system dynamics in~\eqref{eq:system_dynamics} is obtained.

The term $\vf_{ee} \in \mathbb{R}^3$ is an additional virtual force, used as a control term. In its simplest implementation it takes the form of a PD controller:
\begin{equation}
    \bm{f_{ee}} =  \vK(\bm{x_{d}} - \bm{x}) + \vD(\dot{\bm{x}}_{\bm{d}} - \dot{\bm{x}}) \label{eq:impedance_controller}.
\end{equation}
Similar to what is done in inverse dynamics-based impedance controllers, this term is not part of the true robot equations of motion, but modifies them so that the planned control torques produce an additional force in the operational space.
The terms $m, b, k, f_s, x_0, \vv, \vf_{ee}$ are updated in the system dynamics at the beginning of each MPC iteration.


\subsection{Model Identification Adaptive Control (MIAC)} \label{subsec:miac_controller}
\begin{figure}[t]
\includegraphics[scale=0.077]{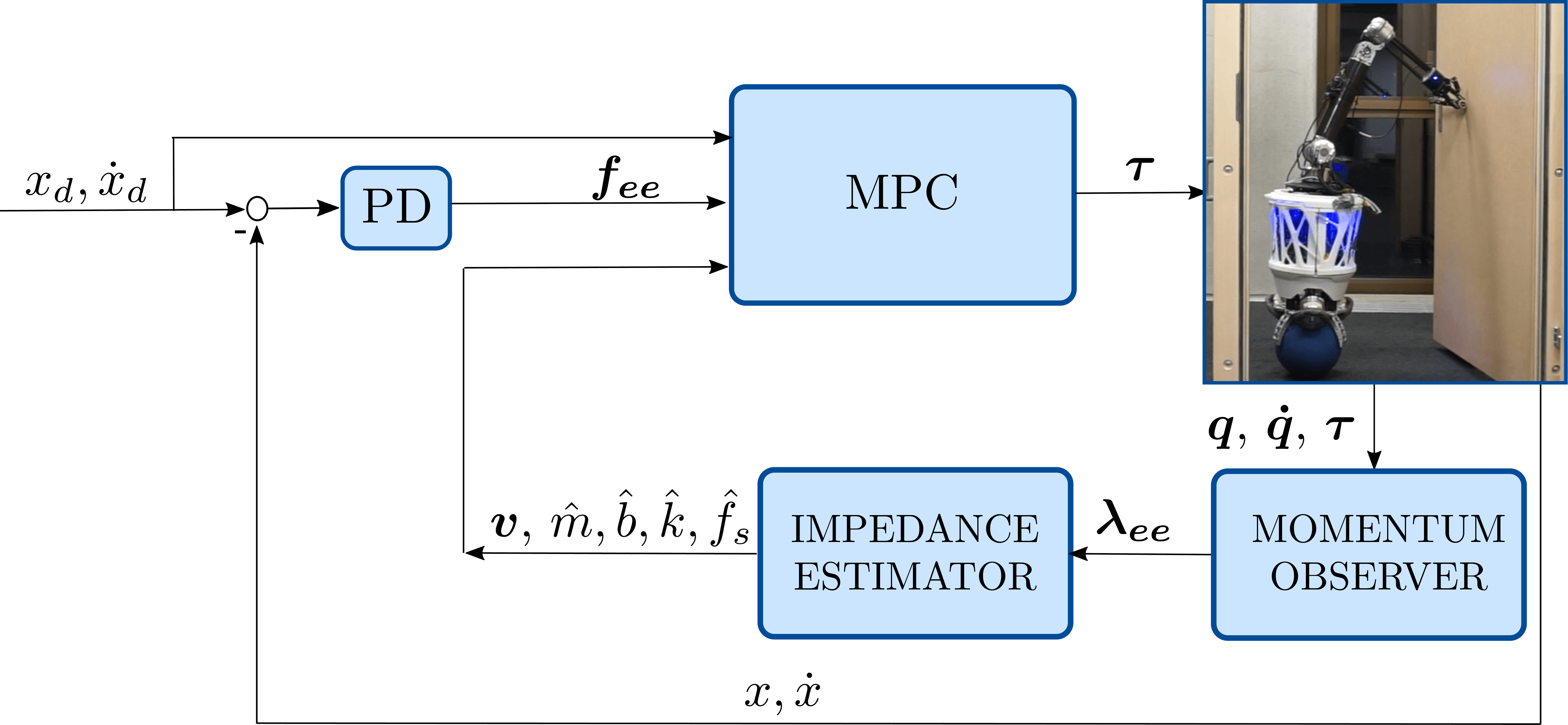}
\caption{Block diagram of the closed-loop system under the MIAC controller.}
  \label{fig:MIAC_block_diagram}
\end{figure}

 The objective of MIAC is to estimate the environment model parameters by means of an online system identification method, and to use these estimates to improve control performance. Since MPC is employed as the main controller module, the estimated parameters can be used in the system dynamics to model the environment. The block diagram of the resulting closed-loop system is indicated in Fig.~\ref{fig:MIAC_block_diagram}. A simpler version of this method would be to directly compensate the external force using measurements from a force sensor. However, estimating the parameters allows for a more accurate prediction by the MPC since a model of the environment is available that can be used during the prediction and optimization over the time horizon. 
 
 Estimating the environment impedance parameters is a classical problem in robotics \cite{love1995environment}. Available methods include Recursive \cite{love1995environment}, Least Mean Square \cite{jasim2015gaussian} estimation, and algebraic manipulations \cite{Becedas2009}. Here, we propose the use of a Kalman filter as online system identification method to estimate the environment's mass $m$, damping $b$, spring stiffness $k$ and static force $f_{s}$. Thus, Eq.~\eqref{eq:environment_eom} is transformed into a state-space model with the substitutions
$
x_{1} :=  x - x_{0}, \quad x_{2} := \dot{x}.
$
Converting the resulting model in discrete time with sampling time $T_s$, the following measurement update equation can be extracted:
\begin{equation}
\small
\lambda[k] = 
\begin{bmatrix}
\frac{x_{2}[k]-x_{2}[k-1]}{T_{s}} &
x_{2}[k] &
x_{1}[k] &
1 
\end{bmatrix}
\begin{bmatrix}
m\\
b \\
k \\
f_{s}
\end{bmatrix}
+
w[k] \label{eq:measurement_update_equation_large},
\end{equation}
with zero mean Gaussian measurement noise $w$.
Eq.~\eqref{eq:measurement_update_equation_large} can be compactly written as:
\begin{equation}
z_{k} = \vH_{k} \vpi_k + w_{k} \label{measurement_update_equation_short},
\end{equation}
with measurement $z$ and estimated parameters $\vpi$. The process model of the filter is chosen as a random walk of the parameters $\vpi$ with process noise $\vn \sim \mathcal{N}(\bm{0},\,\vQ)\,$:
\begin{equation}
\vpi_k =\vpi_{k-1} + \bm{n}. \label{process_model}
\end{equation}
As shown in Eq.~\eqref{eq:measurement_update_equation_large}, the filter measurement is the interaction force $\lambda$ along the manipulation trajectory. This can be acquired with a force sensor, or with an observer approach \cite{haddadin2017robot}. Regarding the measurement update equation \eqref{eq:measurement_update_equation_large}, depending on the task, the two terms $m$ and $f_s$ may be linearly dependent on each other. For instance, if the manipulation is in vertical direction with a free object, then $f_s = mg$. Or, if the manipulated object is pushed in horizontal direction, $f_s = k m g$, where $k$ is a Coulomb friction coefficient. However, to avoid loss of generality, in this formulation they are considered separate from each other. As a result, $m$ is observable only in the presence of persistent accelerations $\ddot x$. 

By providing the estimates $\hat m, \hat b, \hat k, \hat f_s$ to the MPC solver, at the next iteration, the optimal control planner becomes aware of the evolution of the robot-environment model.

\subsection{Model Reference Adaptive Control (MRAC)} \label{subsec:mrac}
\begin{figure*}[!t]
\hspace*{-0.0in} 
   \centering
   \includegraphics[scale=0.09]{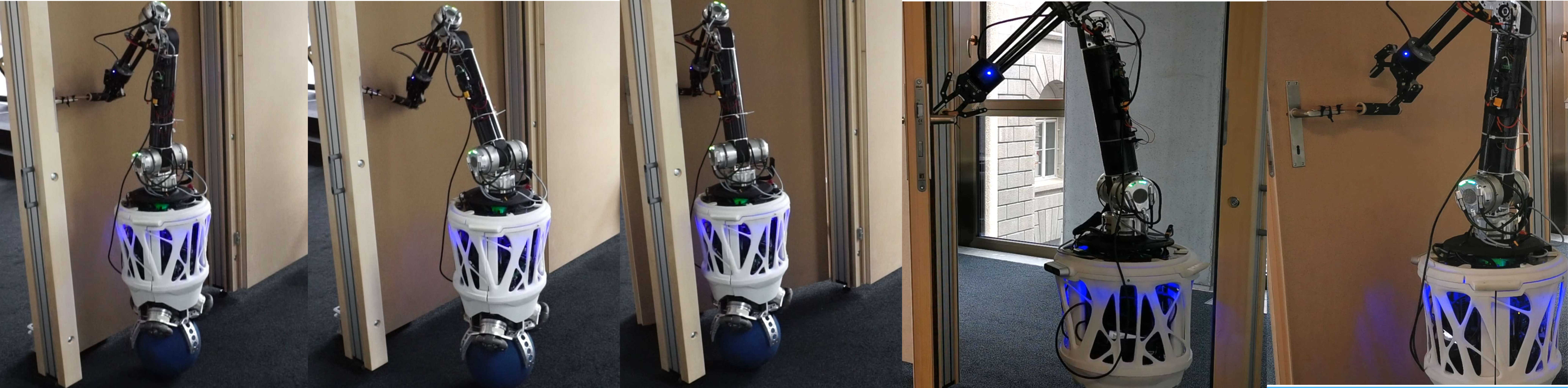}
   \caption{Door opening experiment with a ball-balancing manipulator. The robot is tasked with opening the door up to a desired angle of $70\degree$, while tracking a circular trajectory.}
\label{fig:door_opening_sequence}          
\end{figure*}

As described in \cite{siciliano2010robotics}, \cite{nguyen2018model}, the design method for an MRAC controller consists of finding a control law that is dependent on the reference tracking error and includes adaptive parameters. Then, a Lyapunov function is formulated to derive both the parameters adaption law and a certification of the system stability.
Here, this procedure is simplified since the model \eqref{eq:environment_eom} is linear in the environment parameters. Thus, it can be written as:
\begin{equation}
    m \ddot x + b \dot x + k (x-x_0) + f_s = \vY(\ddot x, \dot x, x)\vpi = \lambda,
\end{equation}
where 
$\vpi:=[m, b, k, f_s ]^T,$
$\vY(\ddot x, \dot x, x) := [
\ddot x, \dot x, x-x_0, 1 
]$.
The control law is defined as
\begin{align}
    \lambda &= \hat m \ddot x_r + \hat b\dot x_r + \hat k(x-x_0) + \hat f_s + k_s \sigma,
    \notag \\
    &= \vY(\ddot x_r, \dot x_r, x)\hat \vpi + k_s \sigma,
    \label{eq:adaptive_control_law}
\end{align}
with $k_s > 0$ as a tuning parameter and with the following definitions:
\begin{align}
&\tilde x = x_d - x,
&&\dot x_r = \dot x_d + \Lambda \tilde x, \quad \Lambda > 0,
\notag \\
&\ddot x_r = \ddot x_d + \Lambda \dot{\tilde x}, 
&&\sigma = \dot x_r - \dot x = \dot{\tilde x}+ \Lambda \tilde x.
\end{align}
A Lyapunov function for the controlled system can be defined as
\begin{equation}
    V(\sigma, \tilde x, \tilde \vpi) = \frac{1}{2}m \sigma^2 + \Lambda k_s \tilde x^2 + \frac{1}{2}\tilde \vpi^T \vK_{\pi}\tilde \vpi > 0,
    \label{eq:mrac_lyapunov_function}
\end{equation}
where $\hat \vpi \in \mathbb{R}^4$ is the vector of adaptive parameters, $\tilde \vpi := \hat \vpi - \vpi$, and $\vK_{\pi} \in \mathbb{R}^{4\times4}$ is a positive definite matrix.
Imposing $\dot V \leq 0$, the adaption law can be derived:
\begin{equation}
    \dot{\hat \vpi} = \vK_{\pi}^{-1}\vY^T(\ddot x_r, \dot x_r, x)\sigma.
    \label{mrac_adaptive_law}
\end{equation}
The control law \eqref{eq:adaptive_control_law}
is the sum of an adaptive term $\vY(\ddot x_r, \dot x_r, x)\hat \vpi$ and a PD-feedback term $k_s \sigma$. The control force to be tracked by the MPC controller is computed as:
\begin{equation}
    \vf_{ee} =  \left(\vY(\ddot x_r, \dot x_r, x)\hat \vpi + k_s \sigma\right)\vv.
    \label{eq:MRAC_controller}
\end{equation}
With respect to the MIAC controller, here the estimates of the environment parameters are not considered in the MPC. The system dynamics thus reduces to Eq.~\eqref{eq:robot_eom} with $\vlambda_{ee} = \bm{0}$. Indeed, the goal of this method is not to exactly compensate the environment model in Eq.~\eqref{eq:environment_eom}, but rather to modify the controller with an adaptive law that guarantees the convergence of the tracking error along the manipulation trajectory. 

\begin{figure}[htp]
\includegraphics[scale=0.07]{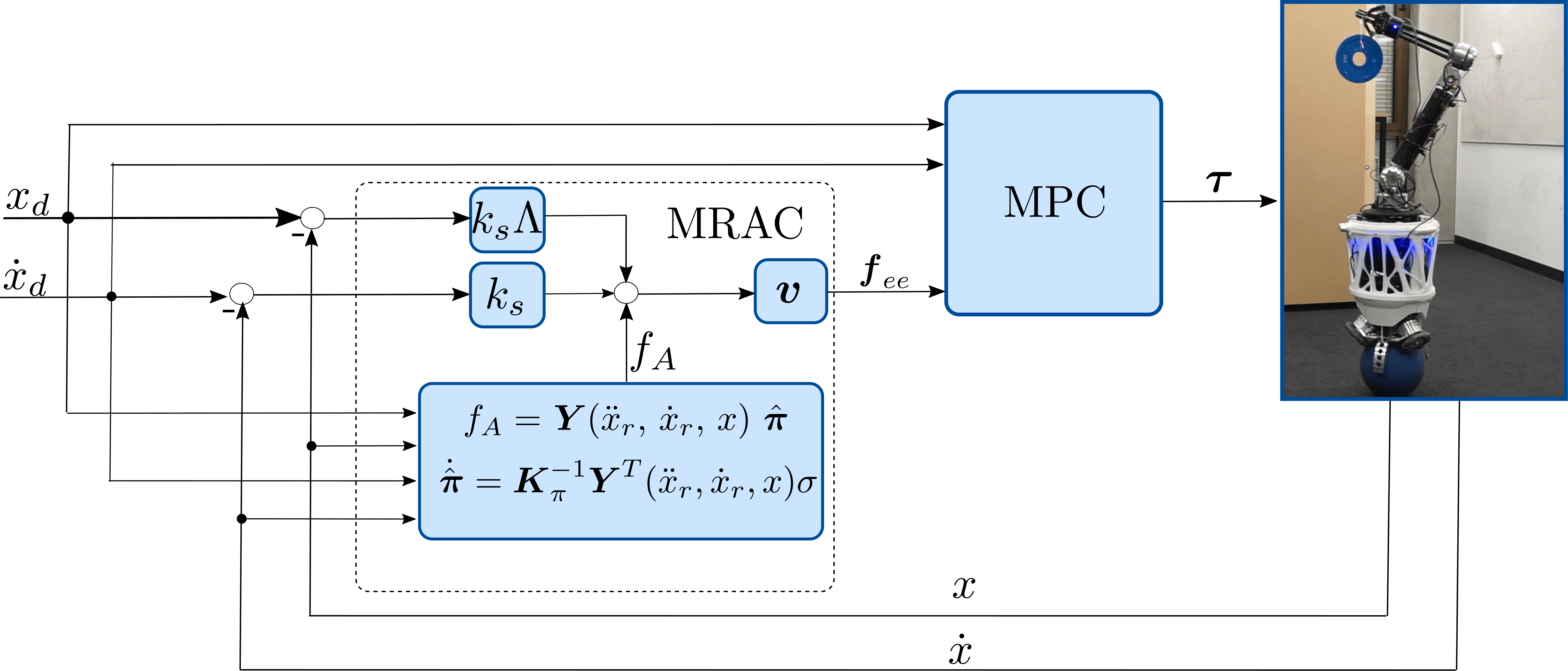}
\caption{Block diagram of the closed-loop system under the cascade MRAC-MPC controller.}
  \label{fig:MRAC_block_diagram}
\end{figure}

\section{EXPERIMENTS} \label{sec:experiments}
The control structures presented in the previous sections \ref{subsec:miac_controller}, \ref{subsec:mrac} are evaluated in two experimental scenarios. In the first scenario, the robot is tasked with opening a hinged door. In the second one, the robot is tasked with lifting an object of unknown mass. The experiments are used to assess how modifying the MPC controller using the environment estimated/adapted parameters can improve the closed-loop tracking performance. A video showcasing the results accompanies this paper \footnote{Available at \url{https://youtu.be/A7_e-UWkfXo}.}.

\subsection{Experimental Setup}
The robot employed for the tests is a ballbot with a 4-degree of freedom (DOF) arm attached to the base and a Robotiq-2F85 gripper as end-effector. The system is shown in Fig.~\ref{fig:ballbot}. It is modeled with 8 DOFs, resulting in a dimensionality of the MPC system of 16 states. The reference torques sent to the actuators are computed as
$\vtau^{ref} = \vtau^{*} + \vK_p(\vq^* - \vq) + \vK_d(\dot \vq^* - \dot \vq,)$, 
where $\vtau^{*}, \vq^*, \dot \vq^*$ are the optimal joint torques, positions and velocities. The MPC solver computes its optimal solution at a frequency of \SI{200}{\hertz} over a time horizon of \SI{1.5}{\second}, and is implemented with the OCS2 toolbox \cite{OCS2}. Rigid-body dynamics and kinematics equations are generated using the RobCoGen library \cite{giftthaler2017automatic}.
State estimation, motion planning and control run on the onboard PC, which is an Intel Core~i7, NUC10i7FNK Mini PC.

The interaction force used as a measurement for the Kalman Filter of Sec.~\ref{subsec:miac_controller} is computed by means of a momentum observer \cite{haddadin2017robot}, which allows to estimate the generalized external torques $\vtau_{\text{ext}}$ from joint position, velocity, and torque measurements. Then, an augmented Jacobian method is employed to retrieve the external force on the end-effector. This is implemented by defining
\begin{equation}
    \vJ_{aug} := \begin{bmatrix}
    \vJ_{base} \\ \vJ_{ee}
    \end{bmatrix},
\end{equation}
where $\vJ_{base} = [\vI_{n_b \times n_b}, \mathbf{0}_{n_b \times (n-n_b)}] $ is the Jacobian corresponding to the $n_b$ degrees of freedom of the base. Then, $\vlambda_{ee}$ is computed from:
\begin{equation}
    \begin{bmatrix}
    \vtau_{\text{base}} \\
    \vlambda_{ee}
    \end{bmatrix} = \vJ_{aug}^{-T}\vtau_{\text{ext}}.
\end{equation}

\subsection{Door Opening}
\label{subsec:door_opening}
In this experiment, the robot should open a door up to a desired angle of $70\degree$. The desired end-effector path is a circular trajectory. 
An extended Kalman Filter (EKF) estimates the door radius, angle, and hinge position, which are used for the re-planning of the time-optimal trajectory generator of Sec.~\ref{subsec:control_task}. Based on given velocity and acceleration bounds and on the current door measurements, the time optimal trajectory generator determines the door angle references and the remaining door-opening time. The parameters of the PD controller \eqref{eq:impedance_controller} are specified in a frame rigidly attached to the door, in order to achieve a higher compliance in the radial direction and a lower compliance in the tangential direction. Since the mass of the original door is not heavy enough to result in a challenging application, it is modified by attaching two boxes of intermediate weight (\SI{1.5}{\kilo\gram}) and larger weight (\SI{4.2}{\kilo\gram}), respectively, behind the door.

During the heavy-door experiment, both the end-effector and the ball-trajectory are commanded to track a desired circular trajectory. The reference for the ball is necessary to avoid a collision with the door frame when the ballbot leans forward while pushing.
Both of the tests are repeated under three different controllers, which are: 1) the baseline MPC controller, with an additional PD force input as in~\eqref{eq:impedance_controller}. 2) the MIAC controller presented in section \ref{subsec:miac_controller}. 3) the MRAC controller presented in section \ref{subsec:mrac}. To compare the tracking performance of the different controllers for the opening of the two doors, we compute the root mean square tracking error (RMSE), and the final error between the desired and estimated door angle which are visible in Table~\ref{table:controller_comparison_both}.

\subsubsection{Light door}
\begin{figure}[t]
\centering
\includegraphics[width=\linewidth]{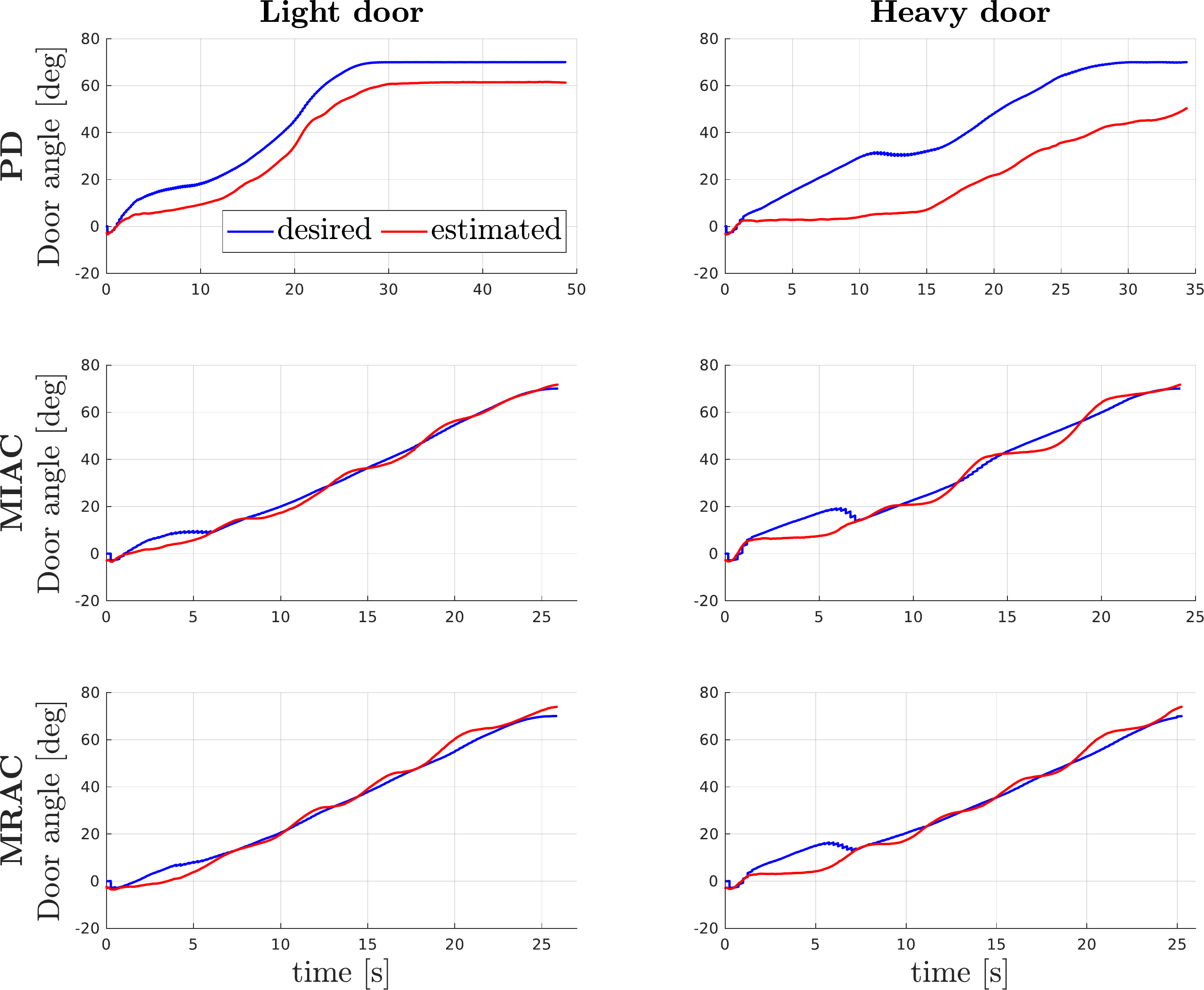}
\caption{From top to bottom: desired and estimated door angle under the PD, MIAC, MRAC controllers, for the door opening experiments. The plots are referred to the light and heavy door experiment on the left and right, respectively.}  \label{fig:door_angle_tracking}
\end{figure}

For this experiment, forces in the range of \SI{10}-\SI{15}{\newton} need to be applied to open the door. The door angle tracking performance is shown in Fig.~\ref{fig:door_angle_tracking}. For the baseline controller, the tracking error is large at all times. On the contrary, the tracking error is comparably small for all the other controllers, with a final door error of less than \SI{4}{\degree}. 

\begin{table}[t]
\centering
\caption{RMSE [deg] and final door errors [deg] of the three interaction controllers in the two door-opening test scenarios.}
\begin{tabular}{lrrrr}
\hline
\multicolumn{1}{c}{} & \multicolumn{2}{c}{Light Door}                             & \multicolumn{2}{c}{Heavy Door}                             \\
Controller           & \multicolumn{1}{c}{RMSE} & \multicolumn{1}{c}{Final Error} & \multicolumn{1}{c}{RMSE} & \multicolumn{1}{c}{Final Error} \\ \hline
Impendance (PD)      & 9.5                     & 8.74                           & 19.5                    & 19.7                          \\
MIAC                 & 2.7                     & \textbf{-1.64}                   & \textbf{4.9}                     & \textbf{-1.7}                    \\
MRAC                 & \textbf{2.6}          & -3.9                            & 5.8            & -3.9                            \\ \hline
\end{tabular}
\label{table:controller_comparison_both}
\end{table}

\begin{figure}[htp]
\centering
\includegraphics[width=\linewidth]{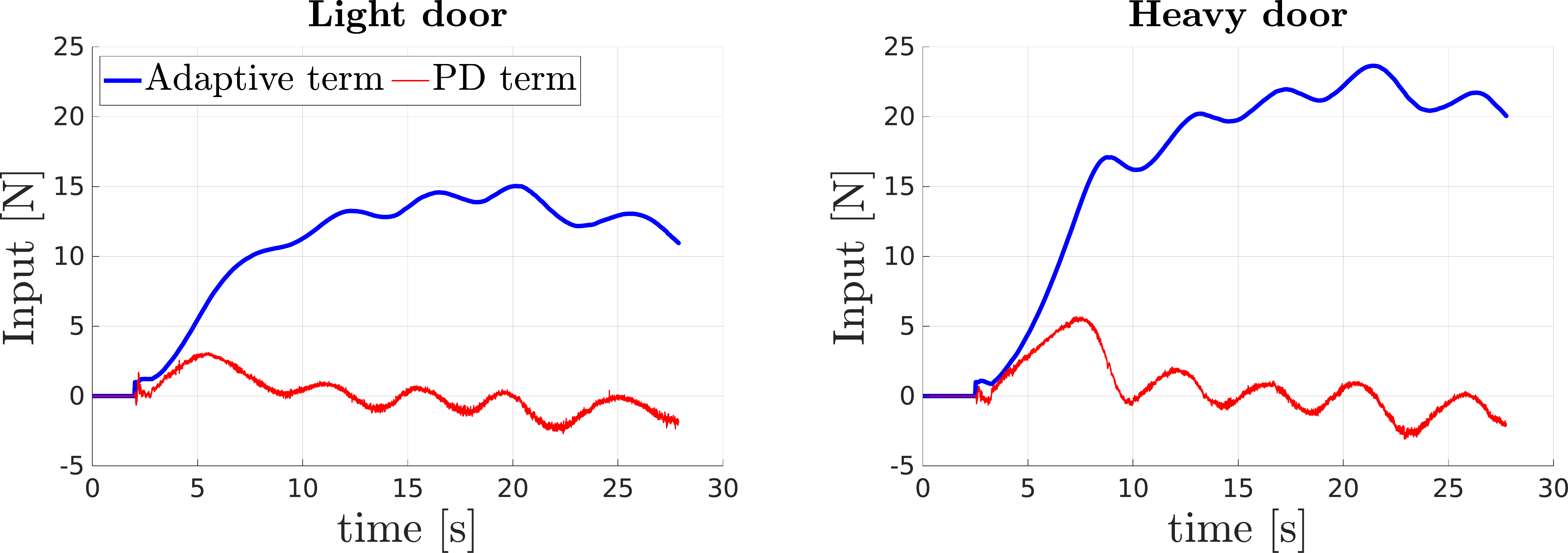}
\caption{Contribution of the adaptive and PD terms in the MRAC control law \eqref{eq:adaptive_control_law} for the light door opening test (on the left) and heavy door opening test (on the right).}
  \label{fig:adaptive_term_contribution}
\end{figure}

\subsubsection{Heavy door}
In this experiment, forces in the range of \SI{20}-\SI{25}{\newton} need to be applied to open the door. The door angle tracking performance for this test is visualized in Fig.~\ref{fig:door_angle_tracking}. We notice that the desired and measured door angles exhibit more oscillations than in the previous test. 
In fact, due to the higher required pushing force, the controller plans to lean the base forward at a larger angle. Tilting the base additionally influences the ball position, resulting in a trade-off between this effect and the door frame collision avoidance term.

\begin{figure}[t]
\centering
\includegraphics[width=\linewidth]{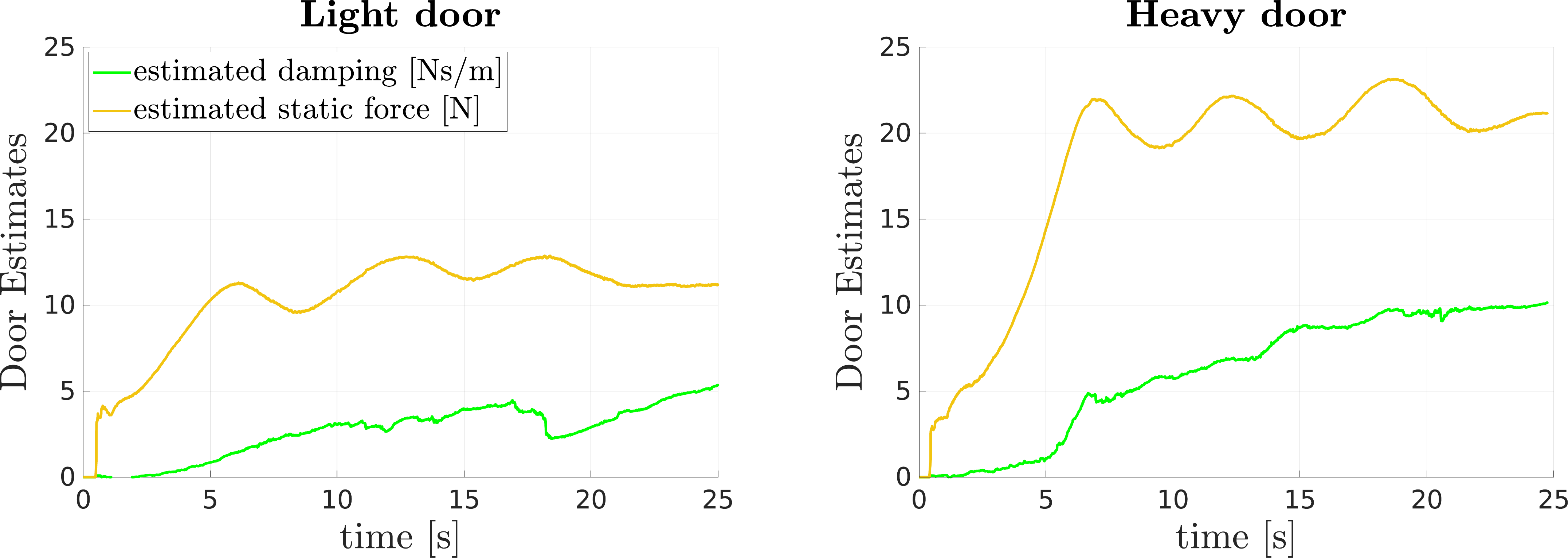}
\caption{Estimation of the damping and static force parameters for the light door opening experiment (on the left), and heavy door opening experiment (on the right) using the MIAC controller.}
  \label{fig:MIAC_estimates}
\end{figure}

Both the MIAC controller and the MRAC-MPC controller produce a considerable reduction of the tracking error, as it can be verified in Table~\ref{table:controller_comparison_both}.  The MRAC controller can fast adapt to the unknown environment due to the balance between the PD and adaptive terms, whose contribution is shown in Fig.~\ref{fig:adaptive_term_contribution}. As the time passes, the PD term becomes less relevant and most of the control action comes from the adaptive input. Regarding the MIAC estimates, since the employed door has no spring behaviour, the most significant parameters are the door damping and static friction, whose estimates are plotted in Fig.~\ref{fig:MIAC_estimates}. 
The non-adaptive baseline impedance controller performs poorly without re-tuning of the stiffness and damping gains and is not able to open the door up to the desired angle. 

\subsection{Lifting Unknown Objects}
\label{subsec:lifting}
In this experiment, the robot is tasked with lifting an unmodeled payload of \SI{2}{\kilo\gram}.
One of the assumptions under which the interaction controllers are derived in section~\ref{sec:interaction_control} is that the robot dynamic model in~\eqref{eq:robot_eom} is perfectly known.
To evaluate how this assumption may influence the controller performance, the same experiment is executed both in a Gazebo simulation, where the robot model is perfect, and on the physical system, where some degree of model mismatch is present. The Gazebo simulation model is based on \cite{minniti2019whole, fankhauser2010modeling}. While in the door opening scenario the robot interacts with a statically stable object, the equilibrium of the object being lifted is attained at a non-zero interaction force, which adds an element of complexity to the task.

\subsubsection{Simulation}

\begin{figure}[t]
\includegraphics[width=\linewidth]{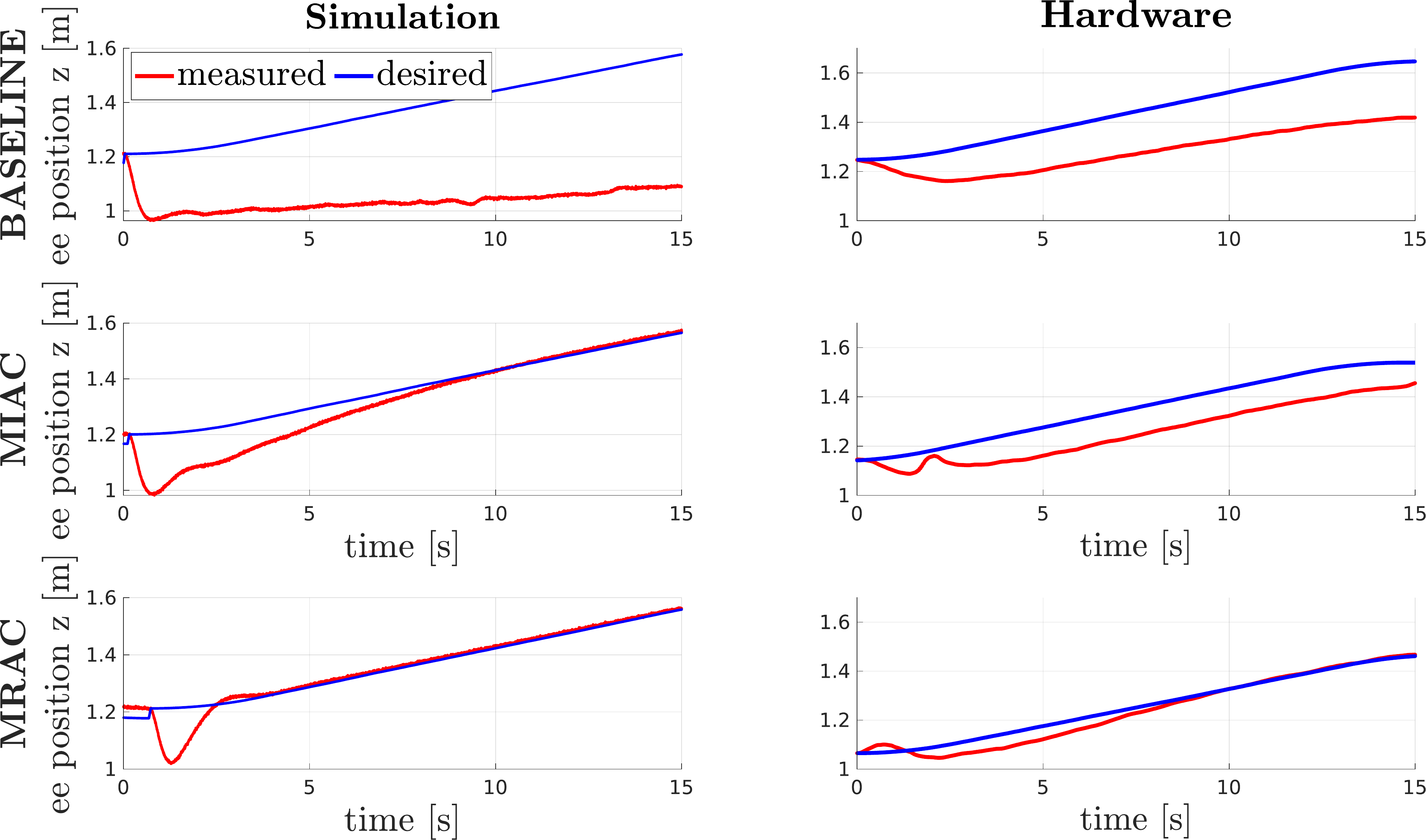}
\caption{From top to bottom: measured and desired end-effector position for the unknown object lifting experiment under the non-adaptive baseline MPC, MIAC, and MRAC, respectively. The plots are referred to simulations (on the left), and hardware tests (on the right).}\label{fig:lifting_tracking}
\end{figure}

The measured and desired trajectory along the end-effector vertical direction is depicted in Fig.~\ref{fig:lifting_tracking}. Analyzing the plots for the simulation test under the baseline controller, it can be inferred that, if the MPC is unaware of the lifted payload, the robot is not able to track the desired trajectory. Conversely, the end-effector position error is minimized if the environment online estimates are included in the optimization, in both an estimated or adaptive manner, as it can be noticed from the plots referring to the MIAC and MRAC controllers.

\subsubsection{Hardware}

The same experiment is performed on the physical system, with corresponding plots shown in Fig.~\ref{fig:lifting_tracking}. The baseline MPC controller has an increasing tracking error. This gets reduced if the whole-body MPC is combined with the proposed interaction control formulation. However, the MIAC controller is strongly influenced by imprecise actuator torque measurements used for the force estimation, and consequent degradation of the environment identification. As a result, the closed-loop system under the MIAC controller has a slower convergence than in the simulation.
On the contrary, the MRAC controller manages to consistently reduce the tracking error in the same amount of time, with the payload being successfully moved to the desired height. Furthermore, we point out that the same gains were used for the MRAC in both the door opening experiment from section \ref{subsec:door_opening} and the lifting test. However, as shown in figure Fig.~\ref{fig:lifting_tracking}, for the MIAC controller, direct transfer to a different task while maintaining the same tracking behavior was not possible with the same parameters used for system identification.

\section{CONCLUSIONS AND FUTURE WORK} \label{conclusion}
This paper focuses on making a whole-body MPC controller adaptive to mobile manipulation tasks in unknown environments. Thus, a system identification and an adaptive control method are proposed to extend the MPC formulation. From experimental tests, it can be concluded that both the MIAC method and the cascade MRAC-MPC outperform the original MPC controller in mobile manipulation tasks applied to unknown environments. However, the MRAC method generalizes better across tasks. In fact, it is independent of interaction forces estimation or the convergence of the MIAC system identification method.
The employed modeling strategy was derived under the assumption that the environment can be described by a linear mass-spring-damper system. 
As a future direction, we intend to generalize the formulation to general multi-DOF tasks (e.g. a collaborative task). Furthermore, so far, the objective has been to achieve the desired closed-loop performance in practice. Providing theoretical guarantees for stability and feasibility in MPC is an active research area, which we will further pursue in the context of robot-environment interaction.

 \bibliographystyle{./bibtex/myIEEEtran} 
 \bibliography{./bibtex/IEEEabrv,bibtex/mybib}

\end{document}